%% file: paper.tex
\documentclass[sigconf]{acmart}
\usepackage[ruled,linesnumbered]{algorithm2e}
\usepackage{array,multirow,graphicx}
\usepackage{multicol}
\newlength{\textfloatsepsave}
\usepackage{soul}
\usepackage{color}
\newcommand{\todo}[1]{{\color{red}#1}}
\newcommand{\vm}[1]{{\color{violet}vm:#1}}

\SetCommentSty{mycommfont}
\usepackage{scalerel}
\DeclareMathOperator*{\redop}{\scalerel*{\odot}{\textstyle\sum}}
\DeclareMathOperator*{\maskop}{\scalerel*{\Theta}{\textstyle\sum}}
\setcopyright{acmcopyright}
\copyrightyear{2020}
\acmYear{2020}
\acmDOI{DOI}
\copyrightyear{2020}
\acmYear{2020}
\setcopyright{acmcopyright}\acmConference[GECCO '20]{Genetic and Evolutionary
Computation Conference}{July 8--12, 2020}{Cancún, Mexico}
\acmBooktitle{Genetic and Evolutionary Computation Conference (GECCO '20), July
8--12, 2020, Cancún, Mexico}
\acmPrice{15.00}
\acmDOI{10.1145/3377930.3390188}
\acmISBN{978-1-4503-7128-5/20/07}

\begin{document}
\title{Semantically-Oriented Mutation Operator in Cartesian Genetic Programming for Evolutionary Circuit Design} 


\newcommand\sharedaffiliation{  
Brno University of Technology, Faculty of Information Technology,\\ IT4Innovations Centre of Excellence\\ Brno, Czech Republic\\
}

\author{David Hodan}
\affiliation{}
\email{ihodan@fit.vutbr.cz}
\author{Vojtech Mrazek}
\orcid{0000-0002-9399-9313}
\affiliation{}
\email{mrazek@fit.vutbr.cz}

\author{Zdenek Vasicek}
\orcid{0000-0002-2279-5217}
\affiliation{}
\email{vasicek@fit.vutbr.cz}

\begin{abstract}
Despite many successful applications, Cartesian Genetic Programming (CGP) suffers from limited scalability, especially when used for evolutionary circuit design.
Considering the multiplier design problem, for example, the 5$\times$5-bit multiplier represents the most complex circuit evolved from a randomly generated initial population.
The efficiency of CGP highly depends on the performance of the point mutation operator, however, this operator is purely stochastic.
This contrasts with the recent developments in Genetic Programming (GP), where advanced informed approaches such as semantic-aware operators are incorporated to improve the search space exploration capability of GP.
In this paper, we propose a semantically-oriented mutation operator (SOMO) suitable for the evolutionary design of combinational circuits.
SOMO uses semantics to determine the best value for each mutated gene.
Compared to the common CGP and its variants as well as the recent versions of Semantic GP, the proposed method converges on common Boolean benchmarks substantially faster while keeping the phenotype size relatively small.
The successfully evolved instances presented in this paper include 10-bit parity, 10+10-bit adder and 5$\times$5-bit multiplier. The most complex circuits were evolved in less than one hour with a single-thread implementation running on a common CPU.
\end{abstract}


\maketitle

\section{Introduction}
The design of combinational logic circuits represents a popular research topic addressed by the evolutionary computing community since the early nineties. The first efforts for evolving various combinational circuits were made by Koza~\cite{Koza:1992:book1}, Coello et al~\cite{Coello:1997} and Miller et al~\cite{Miller:1998}.
Koza investigated the evolutionary design of the even-parity problem in his extensive study related to Genetic Programming (GP). Coello et al. employed the same algorithm to evolve 2-bit adders and multipliers. In contrast to these studies, Miller used a variant of genetic algorithm and successfully evolved more complex circuits such as 4-bit adders and 3-bit multipliers. But the evolutionary design of logic circuits has attracted more attention of researchers with the advent of a new form of genetic programming called Cartesian Genetic Programming (CGP) proposed in 2000~\cite{Miller:cgp}.

Since its introduction, CGP has been adopted by many researchers and successfully applied to many application areas including optimization of Boolean circuits~\cite{Miller:cgp:book,Miller:2019,Vasicek:eurogp15}.
Despite its success, CGP suffers from limited scalability, especially when used for evolutionary circuit design (i.e. when evolving Boolean circuits from scratch)~\cite{VasicekCh:2018}.
To address this issue, researchers introduced various strategies~\cite{vasicek:sekanina:ices14,Goldman:2015,Silva:gam:2019}. Nevertheless, the most complex circuit evolved directly (i.e., without introducing a decomposition) from a randomly generated initial population is a 28-input benchmark~\cite{vasicek:sekanina:ices14}. Considering the common arithmetic circuits, 5$\times$5-bit multiplier and 9+9-bit adder represent the largest problem instances evolved from scratch~\cite{Hrbacek:gecco14}.
In fact, the evolutionary design of multipliers represents a tough problem even for small bit widths.
A multi-thread parallel CGP accelerator running for 4 hours on a cluster of 60 computers was used by Hrbacek et al.~\cite{Hrbacek:gecco14} to obtain a single 5$\times$5-bit multiplier.

Crossover and mutation operators are typically used to create offspring in standard GP. In contrast to that, CGP was designed and is still used with a mutation operator representing the driving force of the evolution~\cite{Miller:2019}.
Despite several attempts to introduce a crossover operator to CGP~\cite{clegg:walker:miller:gecco:2007,slany:sekanina:eurogp:07,Silva:cross:2018}, the standard CGP with mutation operator remains the best 
strategy, especially when considering the design of logic circuits.
One of the possible explanations of the mutation operator's superiority over crossover is that the Boolean domain implies a challenging search space, especially for non-trivial circuit structures such as multipliers~\cite{Miller.2000:Principles2,SekaninaCh:2018}. 

The point and probabilistic mutation are the most used forms of mutation in standard CGP~\cite{Miller:2019}. However, there is a considerable length and positional bias associated with the usage of these operators~\cite{Miller:2019}. 
Moreover, the efficiency of the search deteriorates with increasing the problem size due to the presence of inactive gates.
Since many genes in CGP are redundant, mutations often occur only in the redundant regions, which means that the mutated genotype has the same phenotype as its parent.
Various improvements have been proposed to increase the efficiency of the CGP. 
For example, Single Active Mutation (SAM) operator~\cite{Goldman:2013} was proposed to reduce the wasted objective function evaluations. SAM ensures that in addition to some inactive genes, exactly one active gene is mutated. Moreover, different strategies have been proposed to compensate the positional bias~\cite{Miller:2019}.
Despite these improvements, the mutation operator remains still blind in the sense that mutated genes, as well as the value of the mutated genes, are chosen randomly.
This contrasts with the recent development in GP, where the researchers try to incorporate more advanced informed approaches such as semantic-aware operators to improve the capability of exploring the search space~\cite{Ffrancon:2015,Moraglio:2017,Pawlak:2018}. 
According to our best knowledge, the only approach employing a kind of informed (or biased) mutation operator in CGP is the work of Silva et al.~\cite{Silva:gam:2019}. The authors introduced a guided active mutation (GAM) which aims to reduce the number of evaluations needed to find feasible solutions.
GAM consists of modifying one or more active nodes on the subcircuit from the inputs to the output with the smallest number of correct values when compared to the truth table.
It means that only genes associated with the active nodes are mutated. The mutated genes and their values are chosen randomly.
GAM does not mutate the subgraphs of outputs that already produce a valid response according to the truth table.
When evaluated on Boolean problems, however, GAM achieved low success rates when compared to SAM, and it seems to be necessary to combine both strategies together~\cite{Silva:gam:2019}. 
For single-output problems, it degrades to SAM.


This paper proposes a new mutation operator aiming to reduce the number of evaluations needed to design fully working combinational circuits.
The semantic is used to guide the mutation operator and avoid wasted evaluations.
The point mutation in CGP replaces the mutated genes with randomly chosen (but valid) values.
When a gene associated with input connection is mutated in our case, we try to compute the optimal connection. 
Moreover, a different seeding strategy is used in our case. We always start with a candidate solution consisting of inactive nodes only. Similarly to SAM, only genes associated with the active nodes and program outputs are mutated.
The efficiency of the proposed method is evaluated on standard benchmark problems such as parity circuits and adders as well as on hard benchmark problems, which is the design of combinational multipliers in which the fitness in CGP typically stagnates for many generations without any improvement.
The method is compared to the best CGP-based and GP-based methods available in the literature.

\section{Relevant Work}
\subsection{Semantic Genetic Programming}\label{sec:semgp}
GP operators traditionally work in the syntactic space and manipulate the syntax of parents.
The parents can also be modified based on their semantics.
The semantics of a program can be formally defined in a number of ways. It can be a canonical representation, a description of the behavior based on a logical formalism, or a set of
input-output pairs making up the computed function~\cite{Moraglio:2012}.
In the latter case, we can sample the inputs at random, or we can enumerate all possible input combinations. The exhaustive enumeration is typical, especially for Boolean problems.

Semantic GP is a branch of GP that involves semantic information to improve various aspects of GP.
The semantic, for example, can be used to enforce semantic diversity during the evolutionary process~\cite{Moraglio:2012}. Moreover, we can use semantics to boost the performance of the search space exploration by avoiding returning to points that have already been traversed.
Another possibility is to prevent genetic operators from causing a destructive change in fitness\cite{Beadle:2009}.

In 2012, Moraglio et al proposed Geometric Semantic Genetic Programming (GSGP) which uses specific genetic operators, the so-called geometric semantic operators~\cite{Moraglio:2012}.
On many various symbolic regression and classification problems, it has been shown that GSGP provides statistically better results than a common genetic programming and other machine learning methods in terms of the error score~\cite{Martins:2018}. The reason is that by applying these operators, one can effectively create a unimodal error surface for problems such as symbolic regression. The search process conducted in such a search space is more efficient than in the case of a standard GP. However, geometric semantic operators, by construction, always produce offspring that are larger than their parents, causing a fast growth in the size of the individuals. The growth is linear for mutation and exponential for crossover~\cite{Moraglio:2012}. As a consequence of that, the evolved programs undergo unprecedented growth in their size. This leads to excessive usage of memory and computational power, and also results in non-interpretable solutions~\cite{Castelli:2015}.

Several approaches addressing the problem of the exponential growth of GSGP individuals have been developed, but the problem is still considered unsolved. One branch of the methods is based on simplifying the offspring during the evolution. The other approach is to find new and better versions of the semantic crossover operators~\cite{Moraglio:2012,Martins:2018}. Another possibility is making the algorithm more efficient in terms of memory and computational resources~\cite{Castelli:2015}. 

\subsection{Cartesian Genetic Programming}
CGP grew from a method of evolving digital circuits developed by Miller et al. in 1998~\cite{Miller:1998} to address two issues related to the efficiency of common GP -- poor ability to represent digital circuits, and the presence of bloat.
Compared to tree-based GP, CGP represents the problems using directed acyclic graphs (DAGs) encoded using fixed length structures called netlists (i.e., one-dimensional or two-dimensional grid of nodes).
This representation has many advantages~\cite{Miller:2019}. The graphs can represent many types of computational structures with arbitrary number of outputs. The nodes can be used multiply to create more complex blocks. The fixed-length netlists can contain non-coding (inactive) genes allowing the presence of variable-length phenotypes.

\begin{figure}[t]
\centerline{\includegraphics[width=\columnwidth]{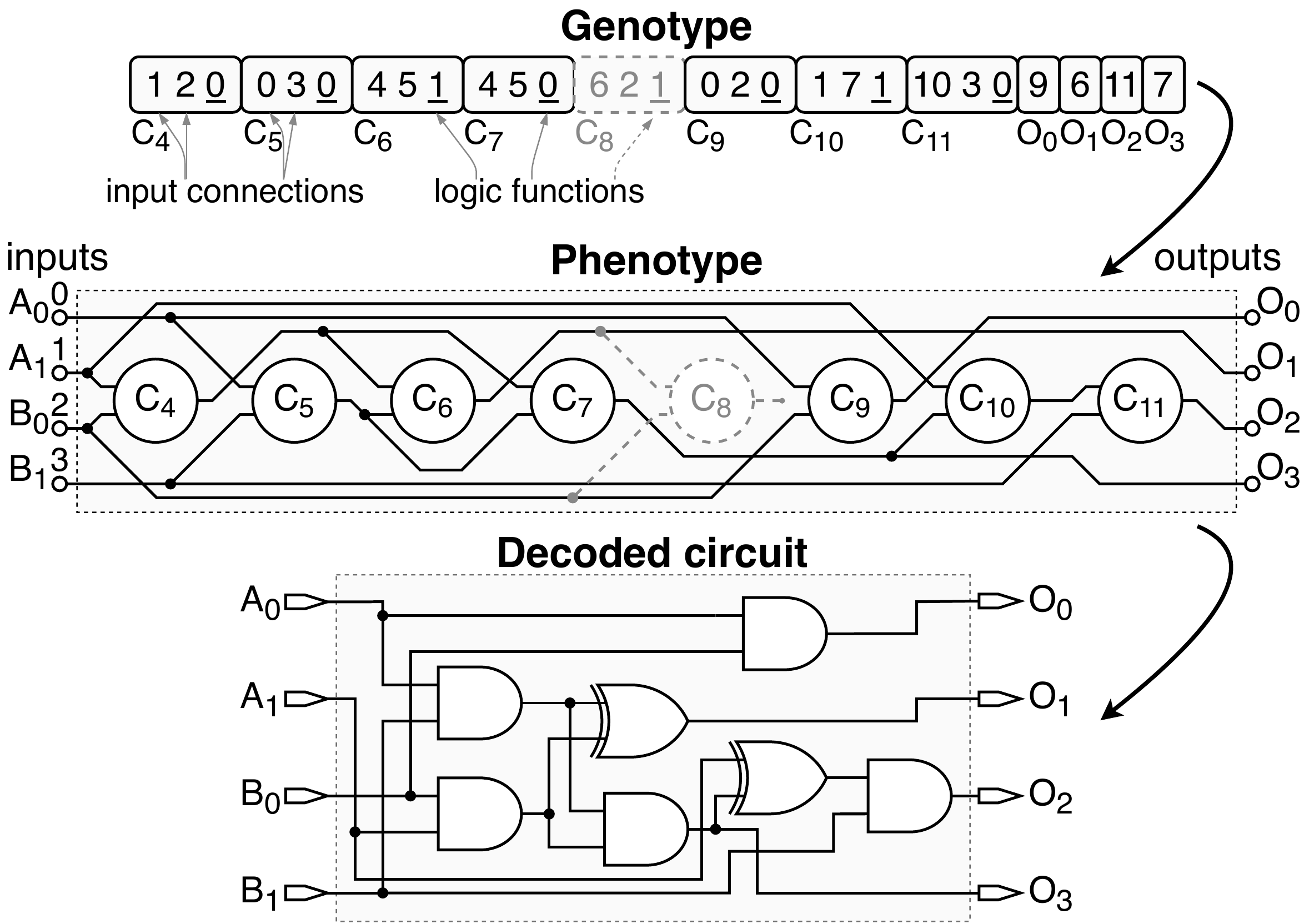}}%
\caption{Example of a CGP individual representing a 2-bit multiplier having four inputs and four outputs encoded by one-dimensional CGP with parameters: $n_i=n_o=4$, $n_c=8$, $n_r=1$, $l=8$, $\Gamma = \{\mathrm{0:and}, \mathrm{1:xor}, \mathrm{2:or}\}$.}\label{fig_cgp}\vspace{-1em}%
\end{figure}

\subsubsection{Circuit Representation}
A two-dimensional array of computational nodes arranged in $n_c$ columns and $n_r$ rows is used to represent the programs (individuals).
CGP utilizes the following encoding scheme.
The program inputs and node outputs are labeled $0, 1, \dots, n_i-1$ and $n_i, n_{i}+1, \dots, n_i+n_c. n_r-1$, respectively, where $n_i$ is the number of program inputs.
Each node input can be connected either to the output of a node placed in previous $l$ columns or to one of the program inputs.
This avoids feedback which is not desirable for combinational circuits.
A candidate solution consisting of two-input computational nodes is represented in the chromosome by $n_c.n_r$ triplets $(c_{in1}, c_{in2}, c_\psi)$ consisting of three genes determining for each node its function $\psi \in \Gamma$, and label of nodes $c_{in1}$ and $c_{in2}$ to which its inputs are connected.
The last part of the chromosome contains $n_o$ genes specifying the labels of nodes or program inputs where the program outputs are connected to.

Depending on the function of a node, some of its inputs may become redundant. For example, a two-input computational node that implements inverter does effectively utilize only one input.
Moreover, some of the nodes may become redundant because they are
not directly or indirectly connected to a program output. 
These nodes are typically referred to as \emph{inactive nodes}.
The presence of inactive nodes enables the existence of neutral mutations. According to many studies, this neutrality is important for an effective search in~CGP~\cite{vassilev:neutral}.
Figure~\ref{fig_cgp} demonstrates the principle of CGP encoding. It can be seen that although eight computational nodes are available in total, not all the nodes have to be employed in the phenotype (resulting circuit). The computational node $C_8$ is inactive so three corresponding genes can be mutated without affecting the phenotype. 

\subsubsection{Search Strategy}
\label{evolution}
CGP employs a ($1 + \lambda$) evolutionary strategy~\cite{Miller:cgp:book}. The algorithm is summarized in Algorithm~\ref{algo:cgp} and consists of the following steps.
The initial population $P$ of the size $1+\lambda$ is generated randomly.
Then, the following steps are repeated until the termination condition is satisfied: (i) the population is evaluated, (ii) a new parent $p$ is selected, (iii) $\lambda$ offspring are created from the parent using mutation operator.

CGP utilizes a point mutation that modifies up to $h$ randomly chosen genes.
The value of a selected gene is replaced with a randomly generated new one.
Note that only certain values can be assigned to each gene to avoid feedback.
The range of valid values depends on the gene position and can be determined in advance.
It is worthwhile to mention that the previous parent is never selected by SelectFittestIndividual procedure as the new parent if there exists at least one individual who obtained the same or better fitness. 
This strategy is important because it ensures diversity between
populations in different generations \cite{Miller:cgp:book}.

When considering an evolution from scratch, the fitness of a candidate solution is typically determined as follows. All the requested assignments are applied to the program inputs, and the number of bits that the candidate solution computes correctly corresponds to the fitness value (additional criteria can be incorporated as well). In case of circuit evolution, this procedure is, in fact, the computation of the Hamming distance (HD) between the specification given in the form of Truth table (TT) and the response of a candidate circuit $p$:
\begin{equation*}
fitness(p) = \mathrm{HD}(p, \mathrm{TT}) = \sum_{\forall x \in \mathbb{B}^{n_i}} \mathrm{OnesCount}(p(x) \oplus \mathrm{TT}(x))
\end{equation*}

The algorithm is terminated when the maximum number of generations is exhausted, or a required solution is obtained, i.e., when $fitness(p) = 0$.
Note that $fitness(\mathrm{NULL})=\infty$. 

\setlength{\textfloatsepsave}{\textfloatsep}\setlength{\textfloatsep}{0pt}%
\begin{algorithm}[t]\SetAlgoLined%
\caption{Standard $(1+\lambda)$-CGP algorithm}\label{algo:cgp}%
create initial population $P=\{p_0,\ldots,p_\lambda\}$; $p \leftarrow \mathrm{NULL}$\; 
\While{terminating condition not satisfied}{
EvaluatePopulation($P$)\;
 $\alpha \leftarrow$ SelectFittestIndividual($P$)\;
  \lIf{fitness($\alpha$) $\leq$ fitness($p$)}{$p \leftarrow \alpha$}
  $P \leftarrow \{p\} \cup \{p'_1=\mathrm{Mutate}(p), \ldots, p'_\lambda=\mathrm{Mutate}(p)\}$\;
}
\Return $p$\;
\end{algorithm}%
\setlength{\textfloatsep}{\textfloatsepsave}%

\section{Proposed Approach}
\subsection{Semantically-Oriented Mutation Operator}
In this work, we propose to replace the purely stochastic mutation operators used in CGP with semantically-oriented mutation operator (SOMO). This work aims to improve the efficiency of the evolutionary design of combinational circuits.
The principle of the proposed operator which replaces the Mutate procedure of Algorithm~\ref{algo:cgp} is summarized in Algorithm~\ref{algo:operator}.

The algorithm accepts the parental solution $p$ and produces its mutated version $p'$.
The mutation operator is operating in phenotype space, hence the first step consists of decoding $p$ and obtaining a DAG formed by a set of nodes denoted as $C$ and a set of edges denoted as $E$.
Every CGP computational node is included in $C$ independently, whether it is active or not. Besides, the program inputs, as well as the program outputs, are treated as nodes. 
To distinguish between program inputs, program outputs, active, and inactive nodes, the set of nodes forming the program inputs $C_{PI}$ and program outputs $C_{PO}$ are provided. Moreover, a mapping $\psi$ assigning a function from $\Gamma$ to each node is produced.
In the next step (line~\ref{algo:operator:getactive}), active nodes are identified. 
The active nodes are those that are directly or indirectly connected to program output.
Then one of the internal active nodes is chosen randomly for mutation.
The mutated node is denoted as $c$.
The mutation can affect either a node function (i.e., $\psi$) or a node connection (i.e., $E$).
Note that only node connection is mutated for a node from $C_{PO}$.
The function of an internal node is mutated with probability $p_f$.
Otherwise, the connection of a single input is mutated.
Finally, the modified DAG is encoded using CGP encoding scheme.
The encoding works as follows. All the nodes are topologically sorted. The unchanged nodes preceding the mutated node $c$ are placed at the beginning of the chromosome, followed by the nodes that have been originally inactive. Finally, the nodes following $c$ together with $c$ are encoded.
This arrangement is not, in fact, necessary from the theoretical point of view. Still, it simplifies the implementation of the mutation operator because we can use the advantage of the CGP encoding and use the connections based on the node indices (i.e., a node can be connected only to those nodes having the lower index to produce acyclic graphs.

\setlength{\textfloatsepsave}{\textfloatsep}\setlength{\textfloatsep}{0pt}%
\begin{algorithm}[t!]\SetAlgoLined
 \caption{Semantically-oriented mutation operator}\label{algo:operator}
 \KwIn{A CGP individual $p$ consisting of $|C|$ nodes}
 \KwOut{A mutated individual $p'$}

 $(C, E, C_{PI}, C_{PO}, \psi) \leftarrow \mathrm{decode}(p)$ \tcp*{decode $p$ as a DAG $(C, E)$ with $C_{PI}$ leaves and $C_{PO}$ roots (outputs); $\psi : C \rightarrow \Gamma$}
 $N \leftarrow \{c \in C\, | \, \exists c_o \in C_{PO}: c \dashv^*_E c_o \}$ \tcp*{get active nodes}\label{algo:operator:getactive}
 $c \leftarrow \mathrm{selectNodeRandomly}(N \setminus C_{PI})$\;
\uIf(\tcp*[h]{mutate node function}){$(\mathrm{rand}(0,1) < p_{f}) \land (c \notin C_{PO})$}{
    $\psi(c) \leftarrow \mathrm{rand}(0,\Gamma-1)$
}\Else(\tcp*[h]{mutate node connection}){
 change connection and function of $p_q$ inactive nodes\;\label{algo:operator:inactive}

 $e \leftarrow \mathrm{selectInputEdgeRandomly}(\{(x,c) \in E | x\in N \})$\;
 $n \leftarrow \mathrm{identifyBestNode}(c,e, (C, E), \psi)$\;
 $E \leftarrow (E \setminus \{e\}) \cup \{(n,c)\}$\;
}
  \Return $p' \leftarrow \mathrm{encode}(C, E, C_{PI}, C_{PO}, \psi)$ \label{algo:operator:encode}
\end{algorithm}%
\setlength{\textfloatsep}{\textfloatsepsave}%

The mutation of a node input connection consists of several steps.
First, some inactive nodes are modified (the amount is defined as ratio $p_q$).
The modification of a node includes the reconnection of all node inputs causing update of $E$ and the change of node function causing update of $\psi$.
The node can be connected to any active node preceding $c$ or any inactive node. New node function is chosen randomly from $\Gamma$.
This step ensures that new genetic material is generated before performing the actual mutation. As a consequence of that, several sub-circuits with provisionally unconnected outputs can arise during this step.
After that, one input of a mutated node $c$ is chosen and reconnected to a node $n$ identified as the most suitable node, i.e., a node whose connection to $c$ causes improvement of the fitness value.

The identification of the most suitable node is based on semantics and is formally defined in Algorithm~\ref{alg:search_connection}. The procedure uses the set of input-output pairs making up the computed function and calculates score for every node of DAG $(C,E)$ that may potentially be connected to the mutated node $c$. Then, the highest-score node is returned.
If more nodes receive the same score, the node closest to the program inputs is preferred.
The score reflects the Hamming distance between values expected at the selected input $e$ of the mutated node $c$ and output values computed by a particular node.
To determine the distance, the algorithm needs to know the set of input-output pairs and outputs for every node, which is obtained by simulating the DAG. We propose to perform the simulation for every possible input combination, which is formally defined as $\mathbb{B}^{n_i}$, but the computation of the score can be in general based on a subset of all possible input combinations.

\setlength{\textfloatsepsave}{\textfloatsep}\setlength{\textfloatsep}{0pt}%
\begin{algorithm}[t!]\caption{Procedure identifyBestNode}\label{alg:search_connection}\SetAlgoLined
 \KwIn{DAG $(C, E)$, node function assignment $\psi$, 
 selected node $c$ and its input $e$, specification in the form of a Truth table $TT(x)$, where $x\in\mathbb{B}^{n_i}$, $\mathbb{B}=\{\mathrm{`0`},\mathrm{`1`}\}$ }
 \KwOut{The most suitable node $n \in C$}

 initialize score(n) to 0 for every $n\in C$\;
 $N \leftarrow \{n \in C | \neg (c \dashv^*_E n) \}$\tcp*{get candidates for connection}
 \ForEach(\tcp*[h]{perform simulation for all inputs}){$x \in \mathbb{B}^{n_i}$}{
   $val \leftarrow$ evaluate $N$ for input $x$ \;

   $val_{e=\mathrm{`0`}} \leftarrow$ evaluate $C\setminus N$ for input $x$ and $e$ forced to '0' \;
   $val_{e=\mathrm{`1`}} \leftarrow$ evaluate $C\setminus N$ for input $x$ and $e$ forced to '1' \;

\tcp{determine desired input value for each output}

$req \leftarrow \redop_{o \in C_{PO}} \left( \maskop\left(TT(x)^{[o]}, val_{e=\mathrm{`0`}}^{[o]}, val_{e=\mathrm{`1`}}^{[o]}\right)\right)$

\tcp{update score of each node}
    \ForEach{$n \in N$}{ \label{alg:search_connection:score}
        $score(n) \leftarrow score(n)  +  \mathrm{HD}^{*}(req, val^{[n]})$\;
     }
  }
  \Return $\mathrm{argmax}_{n \in N} score(n) $\;
\end{algorithm}%
\setlength{\textfloatsep}{\textfloatsepsave}%

Algorithm~\ref{alg:search_connection} starts with determining the set of candidate nodes $N$. This set includes nodes that are not directly or indirectly connected to $c$ and whose connection thus does not cause a cycle.
Then, DAG is simulated for every input combination.
A particular input combination $x$ is applied at the program inputs (i.e., the outputs of DAG leaves $C_{PI}$) and value at the output of every node is determined.
The simulation is divided into three parts. Nodes included in $N$ are simulated at first.
Then, the remaining nodes are simulated with the knowledge of the outputs at nodes included in $N$ for two different cases. The first case reflects the situation when the mutated input $e$ is forced to logic zero, and the second one the situation when $e$ is equal to logic one.
This arrangement helps us to investigate which value the input $e$ should take for a particular combination at program inputs (denoted as $x$) to achieve a match between the value at the program outputs (denoted as $val_{e=0}^{[o]}$ or $val_{e=1}^{[o]}$) and the specification given in form of a Truth table (denoted as $TT(x)^{[o]}$). Note that the term $[o]$ in superscript points to a Boolean value associated with a program output node $o$.
The desired input value is denoted as $req$ and it can be equal to `0`, `1` or `X`, where `X` means that it does not matter what Boolean value the input $e$ takes. This can happen in two cases. One situation is that neither `0` nor `1` presented at $e$ leads to expected output response that matches the specification. The another situation that can happen is that both input values lead to required response and it does not matter which one will be chosen. 
To determine the value of $req$, we use ternary operator $\maskop$ and reduction operator $\redop$ defined as follows:

\begin{multicols}{2}%
\noindent
\begin{equation*}
\maskop(t, v_0, v_1) = \begin{cases}
\mathrm{`X`} & \mathrm{if~} v_0 = v_1 \\
\mathrm{`0`} & \mathrm{if~} v_0 = t \\
\mathrm{`1`} & \mathrm{if~} v_1 = t \\
\end{cases}
\end{equation*}
\begin{equation*}
\redop(a, b) = \begin{cases}
a & \mathrm{when~} a \neq \mathrm{`X`} \\
b & \mathrm{otherwise}\end{cases}
  \end{equation*}
\end{multicols}

\begin{figure}[t]
    \centering
\includegraphics[width=\columnwidth]{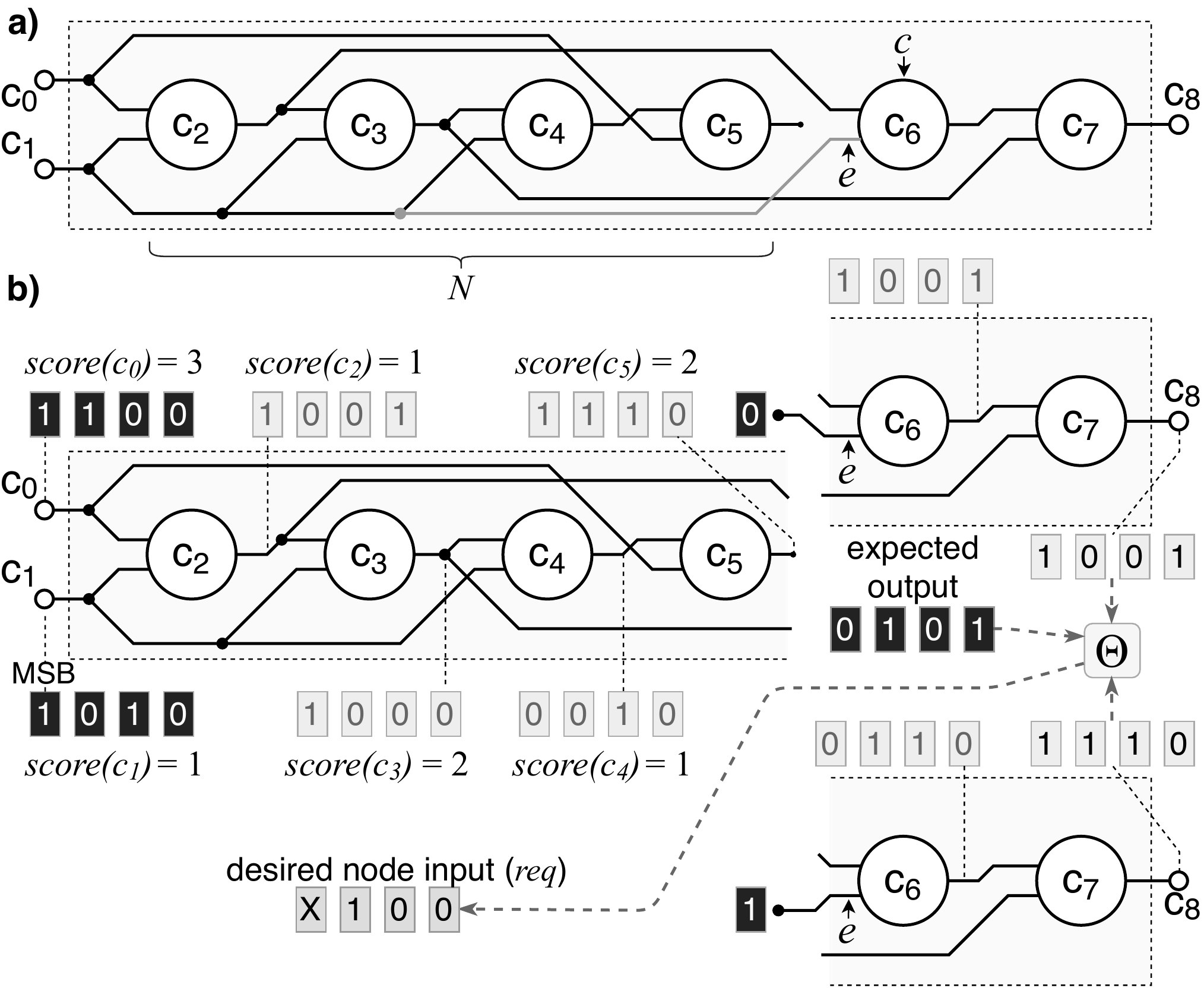}
    \caption{Principle of the most suitable node identification (part b) when mutating a DAG consisting of six computational nodes (part a) given a specification defining the expected output values and mutated node $c_6$ whose second input $e$ is going to be changed. At the end of the process, node $c_0$ is determined as the most suitable candidate.}\label{fig:somoexample}\vspace{-1em}%
\end{figure}

For a particular program output, the operator $\maskop$ takes the required output value according to TT together with the output values obtained when $e=\mathrm{`0`}$ and $e=\mathrm{`1`}$ and determines what input produces the required output. The reduction operator is used to combine the identified values from multiple program outputs to a single value denoted as $req$.
Finally, the score of each candidate node is updated.
In this step, the algorithm compares the value at the node output $val^{[n]}$ which can be either `0` or `1` with $req$ and increments the score provided that a match is detected. The increment corresponds with the inverted Hamming distance which is defined as follows:
\begin{equation*}
\mathrm{HD}^{*}(req, a) = \begin{cases}
1 & \mathrm{when~} (req \neq \mathrm{`X`}) \wedge (req \equiv val) \\
0 & \mathrm{otherwise}\end{cases}
\end{equation*}

The whole process is illustrated in Figure~\ref{fig:somoexample} on a simple problem having two program inputs and single program output.
The DAG consists of six nodes denoted as $c_2, \ldots, c_7$. Node $c_6$ is the mutated node whose second input will be reconnected. Node $c_4$ and $c_5$ are not active and were produced at line \ref{algo:operator:inactive} of Algorithm~\ref{algo:operator}.
We simulate the left part of the DAG denoted as $N$ for every input combination, i.e., $x=(`11`,`10`,`01`,`00`)$. The right part is divided into two separate cases as described above and simulated too.
At the end of the simulation, we receive four responses at the output of every node in $N$.
For node $c_5$, for example, we obtained these values: $val^{[c_5]} = (`1`, `1`, `1`, `0`)$ for the inputs in $x$.
The specification requires the combination $TT^{[c_8]}=(\mathrm{`0`, `1`, `0`, `1`})$ at the program output corresponding to the node $c_8$. We obtained the response $val_{e=\mathrm{`0`}}^{[c_8]}=(`1`, `0`, `0`, `1`)$ and $val_{e=\mathrm{`1`}}^{[c_8]}=(`1`, `1`, `1`, `0`)$ when $e$ is forced to `0` and `1`, respectively.
Looking at the rightmost value, we can see that the specification requires `1`, which is available when $e=\mathrm{`0`}$.
The next required value is available when $e=\mathrm{`1`}$ and the third when $e=\mathrm{`0`}$.
In the last, i.e. leftmost, case neither $e=\mathrm{`0`}$ nor $e=\mathrm{`1`}$ provide `0` at the output, hence the corresponding position of $req$ will be equal to `X`. The complete content of $req$ is equal to $(`X`, `1`, `0`, `0`)$ and forms a kind of pattern which needs to be compared with outputs of all nodes that can potentially be connected at the input of $c_6$.
The highest score is received for $c_0$ because three out of four values corresponds to the pattern. Therefore, $c_0$ is returned as the result of the procedure identifyBestNode.

\subsection{Population Initialization}
Compared to the standard CGP, we propose to use a different strategy for population initialization.
We hypothesize that it is better to start with a candidate solution having no active gate to maximize the efficiency of the proposed approach and minimize the number of active gates of the evolved solutions.
Therefore, the initial population consists of individuals whose program outputs are connected directly to one of the program inputs. The selection of the inputs is done randomly.
As the program outputs are treated as nodes in Algorithm~\ref{algo:operator}, the mutation operator naturally selects one of the output nodes in the first generations.

\subsection{Properties}
\subsubsection{Linear size-dependent overhead against CGP}\label{sec:somooverhead}
Similarly to SAM~\cite{Goldman:2013}, mostly adopted in CGP, a single active node is always mutated.
The mutation of active nodes helps to reduce the wasted evaluations and improves the efficiency of the search.
Considering this fact, the efficiency of the search performed by SOMO expressed in terms of the number of generations needed to find a fully working solution needs to be at least the same as in CGP with SAM.
However, the computational complexity required to create a single generation is higher in SOMO. The overhead compared to the CGP is linear w.r.t. the number of nodes and can be controlled by $p_q$, which determines together with the number of active nodes the total number of simulated nodes. Compared to CGP, we need to simulate a part of the circuit twice. In addition, comparison with values at node outputs is needed in SOMO to calculate score of each node (line~\ref{alg:search_connection:score} in Algorithm~\ref{alg:search_connection}).

\subsubsection{Absence of uncontrollable growth in the size}
As discussed in Section~\ref{sec:semgp}, semantic GP suffers from fast growth in the size of the individuals because the semantic operators always produce offspring that are larger than their parents.
In our case, the growth in the size is linear in the worst case (a constant number of nodes can be activated in each generation).
But the main feature of SOMO is that it can deactivate already active nodes in the course of the evolution.
Moreover, some other advantages of CGP encoding have been implicitly inherited. For example, CGP uses a limited number of computational nodes and naturally does not suffer from a phenomenon called bloat~\cite{Miller:2019}.

\subsubsection{Compensation of positional and length bias}
It has been shown that CGP is naturally biased towards phenotypes of a given size~\cite{goldman:2013length,Goldman:2015}. Typically a small percentage of the available nodes is utilized.
In addition, there is a strong positional bias in CGP causing an increase in the likelihood that nodes close to the inputs will be active~\cite{Miller:2019}.
To compensate both these effects, Reorder operator was introduced by Goldman et al.~\cite{Goldman:2015}.
Both phenomena are naturally mitigated in SOMO. Node reordering is, in fact, performed by the Encode procedure (line \ref{algo:operator:encode} in Alg.~\ref{algo:operator}). Moreover, the number of active nodes may increase naturally due to the construction of SOMO.

\section{Experimental Evaluation}
\subsection{Experimental Setup}
The proposed semantically-oriented operator was implemented in C++ and integrated within a standard CGP algorithm. The application is implemented as a single thread code.
We adopted the principle of parallel simulation introduced in~\cite{vasicek:slany:eurogp12} to maximize the performance of the circuit simulation. Following Hrbacek et al.~\cite{Hrbacek:gecco14}, each candidate solution is transformed into native 256-bit AVX2 instructions.
This arrangement helps us to evaluate circuits having up to 8 inputs in a single pass through the CGP nodes.

The proposed method is evaluated on the evolutionary design of adders, multipliers, and parity circuits. Although the construction of an optimal parity circuit is a straightforward process, parity circuits are considered to be an appropriate benchmark problem within the evolutionary computation community. Usually, a small set of gates (and, or, not) is used. But even if we allow the usage of the xor gates, the design of parity circuits is hard due to the presence of a deceptive landscape, especially for higher bit widths~\cite{VasicekCh:2018}.
The evolutionary design of multipliers represents a hard problem due to the complexity of the multiplication (the multipliers consist of a sequence of adders reducing the partial products to a single output vector).

The evolutionary algorithm uses population of $1+\lambda$ individuals where $\lambda=1$.
This setting enables us to investigate exclusively the impact of the proposed mutation operator.
In each mutation, all inactive nodes are randomly changed, i.e. $p_q=100\%$.
This design choice allows us to investigate the rate of growth in the size.
Only the connections are modified, i.e. $p_f=0$. The function set consists of common binary logic gates $\Gamma^\oplus=\{$not, and, or, xor, nand, nor, xnor$\}$.
To provide a fair comparison with the literature, the parity circuits are also evaluated with a reduced function set $\Gamma^\star=\{$not, and, or, nand, nor$\}$.

As recommended in the literature, we use the one-dimensional form of CGP array consisting of a single row of computational nodes. The number of rows $n_r$ is fixed to 1.
The number of columns $n_r$ needs to be chosen according to the complexity of the addressed problem. Moreover, the standard CGP typically works better under the presence of a reasonable degree of redundancy~\cite{Miller:2019}, i.e. larger genotypes are needed compared to the minimum number of logic gates required to implement a fully functional circuit.
Table~\ref{tab:initialsizes} reports the number of standard gates denoted as $N$ required to implement ripple-carry adder, array multiplier, and parity circuit.
The number of the available CGP nodes is derived from this parameter as $n_c \in \{N, 2N, 5 N, 10 N, 20 N, 50 N, 100 N, 200 N, 500 N, 1000 N\}$.

\setlength{\textfloatsepsave}{\textfloatsep}\setlength{\textfloatsep}{0pt}%
\begin{table}[t]
    \centering
    \caption{The minimal number of gates $N$ required to implement the selected benchmark problems for a given number of inputs $n_{i}$. Parameter $n_{o}$ is the number of outputs. For parity circuits, the number of and/or/not gates is reported.}
    \label{tab:initialsizes}\vspace{-1em}
    \resizebox{\columnwidth}{!}{
    \begin{tabular}{|c c c c c c c c c c|}\hline
         \bf Adder & 2+2 & 3+3 & 4+4 & 5+5 & 6+6 & 7+7 & 8+8 & 9+9 & 10+10 \\\hline
         $n_{i}$ & 4 & 6 & 8 & 10 & 12 & 14 & 16 & 18 & 20\\
         $n_{o}$ & 3 & 4 & 5 & 6 & 7 & 8 & 9 & 10 & 11\\
         $N$ & 7 & 12 & 17 & 22 & 27 & 32 & 37 & 42 & 47 \\\hline
    \end{tabular}}
    \resizebox{\columnwidth}{!}{
    \begin{tabular}{|c c c c c|}\hline
         \bf Multiplier & 2$\times$2 & 3$\times$3 & 4$\times$4 & 5$\times$5 \\\hline
         $n_{i}$ & 4 & 6 & 8 & 10 \\
         $n_{o}$ & 4 & 6 & 8 & 10 \\
         $N$ & 11 & 33 & 67 & 113 \\\hline
    \end{tabular}
    \begin{tabular}{|c c c c c c c c|}\hline
         \bf Parity & 4 & 5 & 6 & 7 & 8 & 9 & 10 \\\hline
         $n_{i}$ & 4 & 5 & 6 & 7 & 8 & 9 & 10 \\
         $n_{o}$ & 1 & 1 & 1 & 1 & 1 & 1 & 1 \\
         $N$ & 9 & 12 & 15 & 18 & 21 & 24 & 27  \\\hline
    \end{tabular}}
\end{table}%
\setlength{\textfloatsep}{\textfloatsepsave}%

In total, 270 configurations (defined by circuit, chromosome size $n_c$ and function set $\Gamma^\oplus$/$\Gamma^\star$) are analyzed and for each configuration, 15 independent runs are executed. The search is terminated either when a fully functional solution is evolved or when the 12-hour limit is elapsed.
As a consequence of the chosen parameter setting ($p_f=0$, $p_q=100\%$), there is a high chance that the search gets stuck in a local optimum when all the available nodes are active.
Therefore, we implemented an additional termination condition to avoid wasted CPU time.
After 15 minutes without any progress, the algorithm is aborted with the result counted as unsuccessful.
All experiments are conducted on Intel\textsuperscript{\textregistered} CPU E5-2630 @ 2.20~GHz.

For each run, it is evaluated how long it takes to find a fully functional solution $p$ exhibiting $fittness(p)=0$ (denoted as \textit{execution time}), how many generations are needed (denoted as \textit{\# generations}) and how many nodes are active in the discovered solution (denoted as \textit{active nodes}). For each configuration we calculated \textit{computational effort} as defined in~\cite{Koza:1992:book1,Miller:cgp:book} and \textit{success rate} (the proportion of runs where a fully functional circuit is found).
The computational effort is calculated for $z=0.99$.

\begin{figure}[t]%
    \centering%
    \includegraphics[width=\columnwidth]{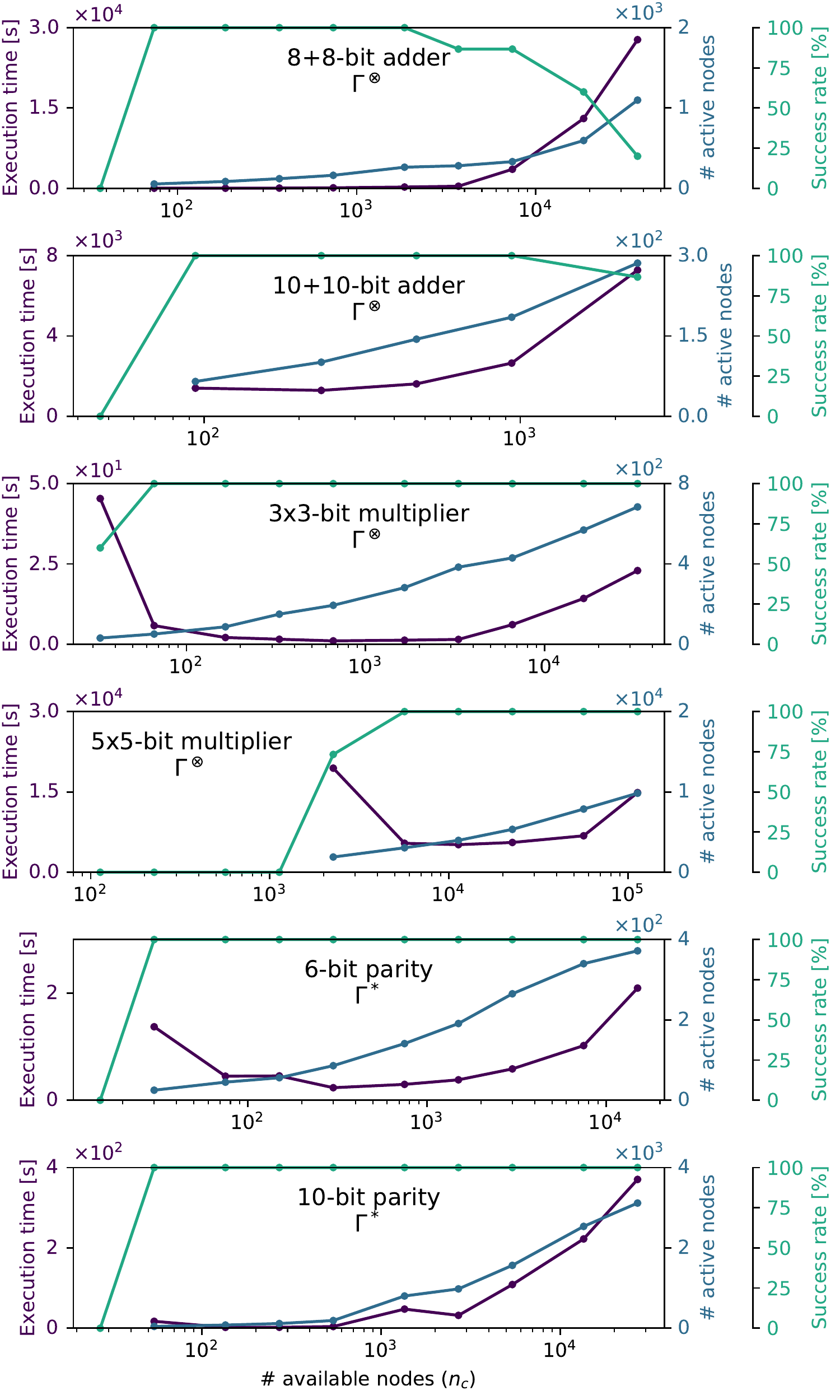}
    \caption{The median execution time needed to find fully functional circuits, the median success rate and the median circuit size for various setting of $n_c$.
    }\label{fig:vana}\vspace{-1.5em}%
\end{figure}%

\subsection{Role of the Available Number of Nodes}
First, we analyze the impact of the amount of available CGP nodes $n_c$ on the efficiency of the search process.
Three aspects are investigated -- the number of generations and the execution time required to evolve a fully functional solution, size of the evolved solutions and the success rate.
The statistics calculated from all 15 runs are shown in Figure~\ref{fig:vana}.
To give a better idea about the real computational complexity, we report the average time needed to find a fully functional solution in this figure.
Due to the limited space, only two instances of each benchmark problem are presented.

As can be seen, the success rate increases with increasing $n_c$.
The results show that it is necessary to introduce some degree of redundancy in $n_c$.
If $n_c$ is equal to $N$, it is hard to find any solution, and the success rate is very low (see 3-bit multiplier) or even zero (adders, 5-bit multiplier, parity circuits).
This is caused by the fact that all the available nodes are active. It is thus hard to make any progress because $p_f=0$.

\begin{figure}[b]
    \centering
\includegraphics[width=\columnwidth]{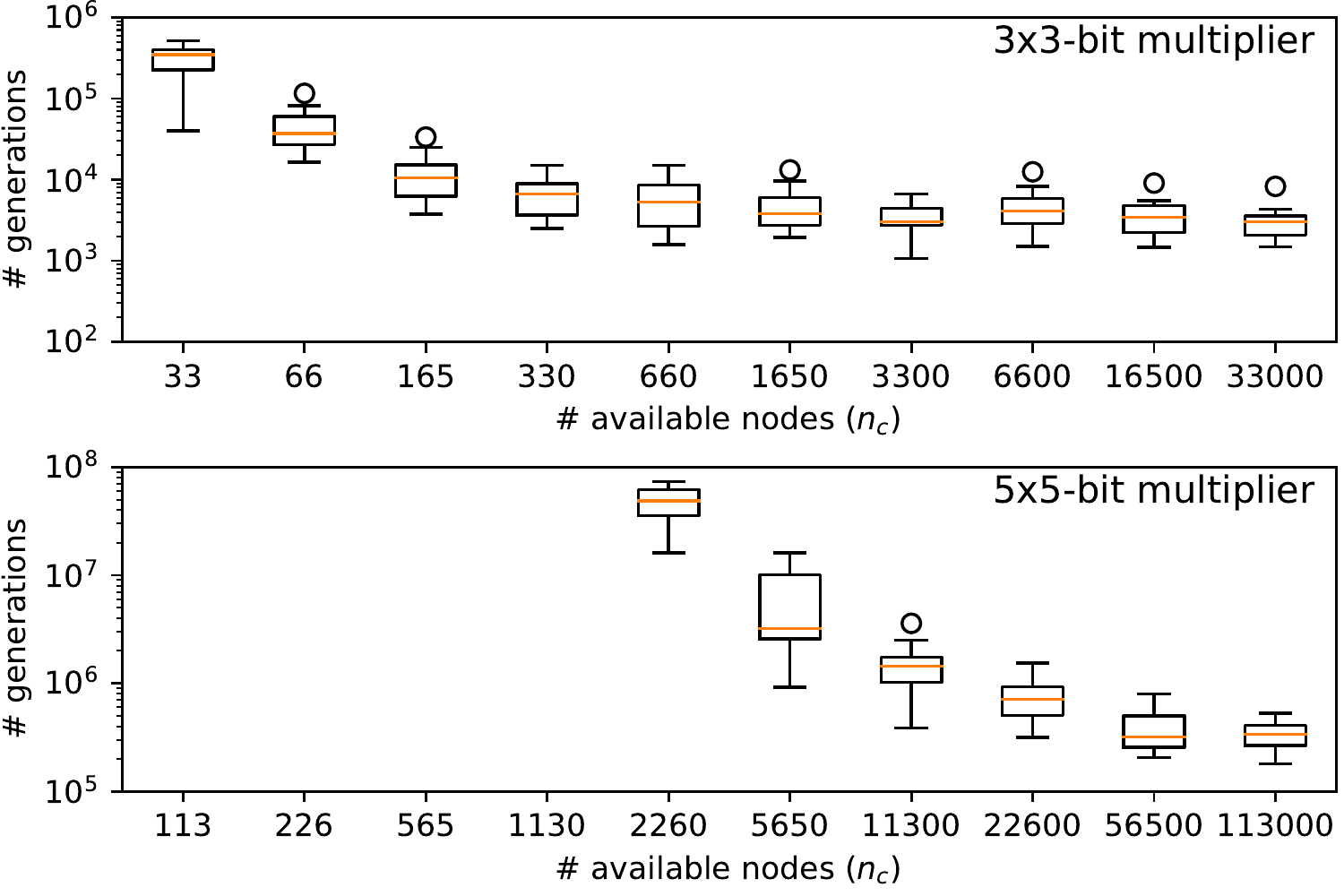}\vspace{-1em}
    \caption{The number of generations required to find a fully functional 3-bit and 5-bit multiplier.}
    \label{fig:params}
\end{figure}

A typical dependency of the execution time on $n_c$ is visible for the 3-bit multiplier.
If $n_c$ is close to $N$, many generations are needed to find a solution. The number of generations decreases with increasing success rate and then remains relatively constant. This effect is visible in Figure~\ref{fig:params}.
The problem is, however, that the execution time increases with increasing $n_c$ because more nodes need to be simulated in each generation. Let us recall that all nodes may be modified (and thus simulated) in every mutation due to the setting of $p_q$ parameter.
This setting was chosen intentionally to see this effect. In a real scenario, it is better to mutate a constant amount of non-active nodes. According to the statistical evaluation, around one hundred nodes were activated at most during a single mutation.

The circuit size grows with the increasing number of available nodes. Despite the exponential growth in $n_c$, the growth in the size is quite linear.
Surprisingly, the evolved circuits are relatively compact even though we did not implement any explicit mechanism which forces the search towards more compact solutions.
For 10-bit parity and $n_c=5N$, for example, we obtained circuits with less than 40 gates even though the number of available nodes is 135.

\begin{figure}[t]
    \centering
\includegraphics[width=\columnwidth]{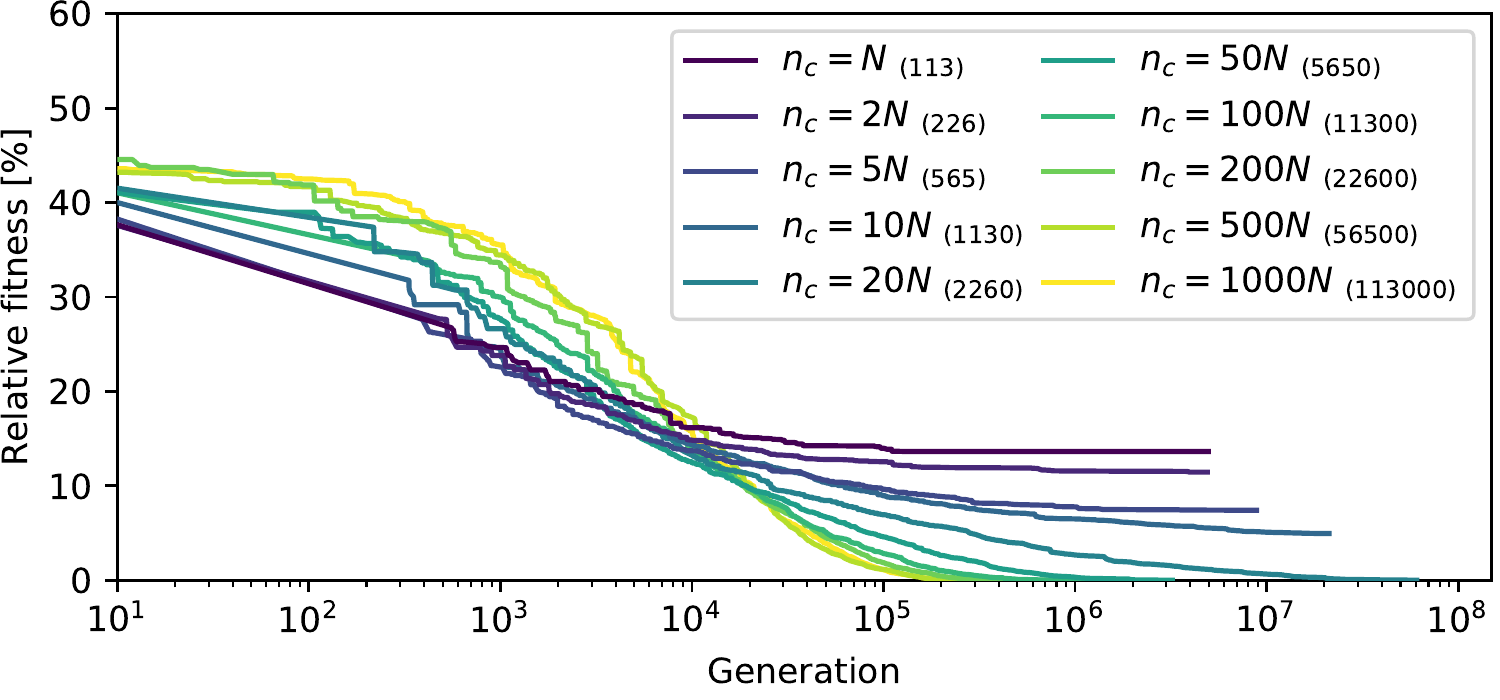}\vspace{-1em}
    \caption{The convergence curves for 5-bit multiplier design and given configuration of $n_c$.
    The median fitness value is reported.
    A fully functional circuit has zero fitness.
    }
    \label{fig:fitness}\vspace{-1.5em}%
\end{figure}

The evolution of the 5-bit multipliers exhibits the worst success rate.
The first four configurations of $n_c$ never reached a solution.
The search typically got stuck at local optima for a longer period and it was prematurely terminated due to the presence of a hard time limit.
When we removed the limit, we were able to evolve multipliers for $n_c=10N$.
Despite that, no multiplier was evolved for lower values of $n_c$.
The explanation can be seen when we look at the results shown in Figure~\ref{fig:params}.
Around 60 times more generations are needed to evolve 3-bit multiplier for $n_c=N$ compared to the setting $n_c=10N$.
It means that the insufficient number of generations is provided for 5-bit multipliers.
Figure~\ref{fig:fitness} shows the typical convergence curves for all configurations.
The higher $n_c$, the faster convergence.

Minimum, maximum and the median number of generations needed to find a fully functional 3-bit and 5-bit multiplier is analysed in Figure~\ref{fig:params}.
We can see that the performance of the proposed method is relatively robust regarding parameter setting. This property is visible on 3-bit multipliers and is also observable on adders and parity circuits where we can see a relative low sensitivity of the computational complexity to the chosen value $n_c$.


\begin{table*}[t]
    \centering
    \caption{Overall statistics computed for a configuration exhibiting the best mean time required to find a working solution
    }\vspace{-1em}
    \label{tab:results}
    \resizebox{\textwidth}{!}{
    \begin{tabular}{|cc|c|c|c||rr|rrlr|rrlr|rrlr|}\hline
       \multicolumn{2}{|c|}{\multirow{2}{*}{\bf Circuit}} & \multirow{2}{*}{$n_{i}$} &  \multirow{2}{*}{$n_{o}$} & \multirow{2}{*}{$n_c$} & \multicolumn{1}{c}{\bf Succ.} & \multicolumn{1}{c|}{\bf Comp.} & \multicolumn{4}{c|}{\bf Execution time [s]} & \multicolumn{4}{c|}{\bf \# Generations (evaluations)} & \multicolumn{4}{c|}{\bf \# Active nodes (circuit size)}  \\
        & &   &  & & \multicolumn{1}{c}{\bf rate} & \multicolumn{1}{c|}{\bf effort} & \it min & \multicolumn{2}{c}{\it mean} & \it max & \it min & \multicolumn{2}{c}{\it mean} & \it max  & \it min & \multicolumn{2}{c}{\it mean} & \it max  \\\hline
        \input{res.tex}\hline
    \end{tabular}
    } 
\end{table*}

\subsection{Overall results}\label{sec:results:overall}
The overall results are summarized in Table~\ref{tab:results}. For each circuit, the statistics related to the execution time, required number of evaluations, and the number of nodes is provided for such $n_c$ that achieves the best mean search time.
The minimal, maximal, and mean values are reported together with the 0.95 confidence level.

We can see that the execution time increases exponentially with the increasing bit width, however, it is still acceptable even for the largest circuits. The number of active nodes is enormous considering the conventional circuit implementations, but the circuit size is not subject of the evolution.
For parity circuits, we report the results for the complete as well as reduced function set.
The reduced function set causes a $20\times$ increase in the execution time and produces approximately $2\times$ larger circuits. 


\subsection{Comparison with the literature}

\subsubsection{Evolutionary circuit design using CGP}
The 6-bit adders were successfully evolved using an advanced mutation operator GAM in~\cite{Silva:gam:2019}. 
More than $10^6$ evaluations were needed for the best configuration (1+3) producing circuits having from 116 to 175 gates. SOMO requires 127x fewer evaluations on average and produces more compact circuits (73 - 150 gates).
As discussed in Section~\ref{sec:somooverhead}, a single SOMO evaluation corresponds up to two truth table evaluations in common CGP.
Considering this fact, SOMO still performs substantially better than GAM.

The computational effort of different CGP mutation and crossover operators is evaluated in \cite{Silva:cross:2018}. Although many powerful techniques improving CGP for digital circuit design have been proposed, the SOMO achieves 30 - 114x lower computational effort compared to the best method (see Table~\ref{tab:soa}). Note that adders with carry input are considered in this comparison. 

\begin{table}[b]\vspace{-1em}%
    \centering%
    \caption{Computational effort of different variants of CGP {\normalfont\small (results for CGP, MC-CGP, MC-ECGP, and X-CGP taken from \cite{Silva:cross:2018})}\vspace{-1em}}\label{tab:soa}%
\resizebox{\columnwidth}{!}{
    \begin{tabular}{|c|rrrr||r|r|}\hline
Circuit    &    CGP    &    MC-CGP &    MC-ECGP    &    X-CGP    &    Our &    Reduction\\\hline
Adder 3+2    &     496,200   & 140,800    &    \bf 24,200    &     54,661    & \bf  798    &    $30 \times$\\
Adder 4+3    &     8,190,400  &     1,286,000    &     1,230,400    &    \bf 360,746    &    \bf  3,171    &    $114 \times$\\
Mult. 2x2    &     52,000   &    \bf 11,200    &     22,400    &     32,962    &    \bf 287    &    $39 \times$\\
Mult. 3x3    &     18,509,600   &     873,600    &    \bf 867,600    &     950,374 &    \bf  15,134    &    $57 \times$\\
\hline
    \end{tabular}
    }
\end{table}

The most complex arithmetic circuits evolved using CGP were reported by Hrbacek et al.~\cite{Hrbacek:gecco14}.
Their multi-threaded parallel implementation of CGP discovered a 5-bit multiplier on a supercomputer cluster in 548 core-hours.
Due to high computational requirements, only a single run was executed.
SOMO can design this multiplier in 42 minutes on average (i.e., 771x faster) on a standard CPU. The 9-bit adders were discovered in 2.3 core-hours compared to the average 170 seconds (50x speedup) needed by SOMO.


\subsubsection{Evolutionary circuit design using Semantic GP}
Ffrancon et al. reports results for parity circuits with the reduced function set $\Gamma^\star$~\cite{Ffrancon:2015}. The authors were able to design 6-, 8-, and 9-bit parity in 164, 622, and 5850 seconds on average, respectively. SOMO accelerates the design process 119x, 586x, and 5098x, respectively. The experiments were conducted on a comparable CPU. However, it is fair to say that the authors implemented the algorithm in Python. The obtained circuits were quite bulky (435, 1972, and 4066 gates) compared to SOMO (75, 106 and 75 gates).

Parity circuits were also evolved by Pawlak and Krawiec using Geometric Semantic GP~\cite{Pawlak:2018}. 6-bit, 7-bit, and 8-bit parity circuits were found with \textsc{RTsSgxm} operator after $100$ generations using a population of thousand individuals (i.e., $10^5$ evaluations).
The average reported circuit size was 298 -- 331 gates (3.7x larger than in our case).
On average, SOMO requires 5772 evaluations to find a solution for 8-bit parity.

\section{Conclusions}
We proposed semantically-oriented mutation operator and took a first step towards a more advanced mutation in CGP. 
The obtained results clearly indicate that the use of the semantically oriented operator is beneficial and significantly improves the search performance of CGP when applied to evolutionary design of combinational circuits.

In our future work, we would like to also use the semantics during mutation of the node functions. This improvement enables us to create a unimodal fitness landscape. In addition, we would like to apply the idea of semantic CGP via semantic mutation to non-Boolean problems.


\begin{acks}
This work was supported by The Ministry of Education, Youth and Sports of the Czech Republic from the National Programme of Sustainability (NPU II); project IT4Innovations excellence in science-LQ1602.
\end{acks}

\clearpage
\bibliographystyle{ACM-Reference-Format}
\bibliography{paper}

\end{document}

%% file: res.tex
\multirow{6}{*}{\shortstack{\bf Adder\\$\Gamma^\oplus$}}	&	2+2	&	4	&	3	&	350	&	100\%	&	365	&	0.01	&	0.04	&	\footnotesize{$\pm 0.01$}	&	0.07	&	66	&	166.9	&	\footnotesize{$\pm 52.7$}	&	365	&	36	&	64.5	&	\footnotesize{$\pm 8.1$}	&	102\\
	&	4+4	&	8	&	5	&	85	&	100\%	&	5,972	&	0.16	&	0.44	&	\footnotesize{$\pm 0.13$}	&	1.12	&	830	&	2,376	&	\footnotesize{$\pm 744$}	&	5,972	&	32	&	41.2	&	\footnotesize{$\pm 2.9$}	&	48\\
	&	6+6	&	12	&	7	&	270	&	100\%	&	20,461	&	0.64	&	2.25	&	\footnotesize{$\pm 0.79$}	&	4.93	&	2,381	&	8,466	&	\footnotesize{$\pm 3,058$}	&	20,461	&	73	&	97.7	&	\footnotesize{$\pm 12.9$}	&	150\\
	&	8+8	&	16	&	9	&	185	&	100\%	&	29,922	&	5.39	&	20.82	&	\footnotesize{$\pm 5.42$}	&	37.36	&	4,074	&	13,872	&	\footnotesize{$\pm 3,665$}	&	29,922	&	76	&	85.9	&	\footnotesize{$\pm 4.3$}	&	102\\
	&	9+9	&	18	&	10	&	210	&	100\%	&	31,206	&	60.69	&	169.99	&	\footnotesize{$\pm 56.27$}	&	412.08	&	6,041	&	14,801	&	\footnotesize{$\pm 4,267$}	&	31,206	&	72	&	90.8	&	\footnotesize{$\pm 5.4$}	&	114\\
	&	10+10	&	20	&	11	&	235	&	100\%	&	42,739	&	532.77	&	1,326	&	\footnotesize{$\pm 383$}	&	2,384	&	9,300	&	23,544	&	\footnotesize{$\pm 6,158$}	&	42,739	&	78	&	101.8	&	\footnotesize{$\pm 6.3$}	&	124\\
\hline \multirow{4}{*}{\shortstack{\bf Mult.\\$\Gamma^\oplus$}}	&	2x2	&	4	&	4	&	220	&	100\%	&	287	&	0.01	&	0.03	&	\footnotesize{$\pm 0.007$}	&	0.06	&	58&164.0	&	\footnotesize{$\pm 38.9$}	&	287	&	33	&	52.3	&	\footnotesize{$\pm 4.5$}	&	64\\
	&	3x3	&	6	&	6	&	660	&	100\%	&	15,134	&	0.33	&	1.26	&	\footnotesize{$\pm 0.47$}	&	3.21	&	1,576	&	5,977	&	\footnotesize{$\pm 2,184$}	&	15,134	&	141	&	196.8	&	\footnotesize{$\pm 14.6$}	&	244\\
	&	4x4	&	8	&	8	&	3,350	&	100\%	&	188,710	&	11.83	&	38.59	&	\footnotesize{$\pm 11.53$}	&	97.21	&	28,743	&	85,253	&	\footnotesize{$\pm 24,848$}	&	188,710	&	773	&	1,049	&	\footnotesize{$\pm 82$}	&	1,328\\
	&	5x5	&	10	&	10	&	11,300	&	27\%	&	742,497	&	1,278	&	2,556	&	\footnotesize{$\pm 1,495$}	&	3,493	&	387,860	&	712,077	&	\footnotesize{$\pm 465,540$}	&	1,090,909	&	3,671	&	4,476	&	\footnotesize{$\pm 1,012$}	&	5,189\\
\hline \multirow{7}{*}{\shortstack{\bf Parity\\$\Gamma^\oplus$}}
	&	5	&	5	&	1	&	60	&	100\%	&	147	&	0.002	&	0.009	&	\footnotesize{$\pm 0.004$}	&	0.03	&	5	&	38.9	&	\footnotesize{$\pm 21.6$}	&	147	&	6	&	18.1	&	\footnotesize{$\pm 4.3$}	&	30\\
	&	6	&	6	&	1	&	75	&	100\%	&	235	&	0.005	&	0.02	&	\footnotesize{$\pm 0.006$}	&	0.04	&	23	&	87.1	&	\footnotesize{$\pm 35.1$}	&	235	&	11	&	23.7	&	\footnotesize{$\pm 6.0$}	&	51\\
	&	7	&	7	&	1	&	90	&	100\%	&	450	&	0.007	&	0.04	&	\footnotesize{$\pm 0.01$}	&	0.07	&	28	&	191.8	&	\footnotesize{$\pm 69.9$}	&	450	&	12	&	28.1	&	\footnotesize{$\pm 6.8$}	&	50\\
	&	8	&	8	&	1	&	105	&	100\%	&	705	&	0.007	&	0.04	&	\footnotesize{$\pm 0.02$}	&	0.13	&	33	&	185.9	&	\footnotesize{$\pm 107.5$}	&	705	&	13	&	33.7	&	\footnotesize{$\pm 5.9$}	&	53\\
	&	9	&	9	&	1	&	240	&	100\%	&	1,029	&	0.009	&	0.05	&	\footnotesize{$\pm 0.03$}	&	0.19	&	43	&	280.5	&	\footnotesize{$\pm 137.9$}	&	1,029	&	47	&	71.1	&	\footnotesize{$\pm 12.8$}	&	119\\
	&	10	&	10	&	1	&	270	&	100\%	&	1,794	&	0.02	&	0.12	&	\footnotesize{$\pm 0.05$}	&	0.35	&	77	&	596.5	&	\footnotesize{$\pm 281.3$}	&	1,794	&	43	&	79.6	&	\footnotesize{$\pm 12.6$}	&	125\\

\hline\hline

\multirow{7}{*}{\shortstack{\bf Parity\\$\Gamma^\star$}}	
	&	5	&	5	&	1	&	240	&	100\%	&	1,636	&	0.06	&	0.13	&	\footnotesize{$\pm 0.05$}	&	0.34	&	264	&	616.1	&	\footnotesize{$\pm 217.4$}	&	1,636	&	61	&	70.6	&	\footnotesize{$\pm 3.8$}	&	86\\
	&	6	&	6	&	1	&	300	&	100\%	&	2,991	&	0.13	&	0.28	&	\footnotesize{$\pm 0.07$}	&	0.56	&	591	&	1,376	&	\footnotesize{$\pm 393$}	&	2,991	&	52	&	84.1	&	\footnotesize{$\pm 12.0$}	&	131\\
	&	7	&	7	&	1	&	180	&	100\%	&	9,406	&	0.19	&	0.61	&	\footnotesize{$\pm 0.30$}	&	1.92	&	1,019	&	3,132	&	\footnotesize{$\pm 1,449$}	&	9,406	&	55	&	74.9	&	\footnotesize{$\pm 7.3$}	&	100\\
	&	8	&	8	&	1	&	210	&	100\%	&	13,078	&	0.38	&	1.13	&	\footnotesize{$\pm 0.38$}	&	2.42	&	2,045	&	5,772	&	\footnotesize{$\pm 1,910$}	&	13,078	&	58	&	85.0	&	\footnotesize{$\pm 8.5$}	&	109\\
	&	9	&	9	&	1	&	240	&	100\%	&	14,102	&	0.61	&	1.59	&	\footnotesize{$\pm 0.35$}	&	2.65	&	2,872	&	7,793	&	\footnotesize{$\pm 1,741$}	&	14,102	&	78	&	106.5	&	\footnotesize{$\pm 11.8$}	&	151\\
	&	10	&	10	&	1	&	135	&	100\%	&	18,950	&	0.67	&	2.36	&	\footnotesize{$\pm 0.56$}	&	3.71	&	3,544	&	11,647	&	\footnotesize{$\pm 2,662$}	&	18,950	&	63	&	78.1	&	\footnotesize{$\pm 5.6$}	&	98\\ 